\pdfoutput=1 
\documentclass{article}
\usepackage{spconf,amsmath,graphicx}
\usepackage{tabularx,booktabs}
\usepackage{lipsum}

\title{Robust Learning at Noisy Labeled Medical Images:\\
	 Applied to Skin Lesion Classification}

 \name{Cheng Xue$^1$ \qquad Qi Dou$^2$ \qquad Xueying Shi$^1$ \qquad Hao Chen$^{1,3}$ \qquad Pheng-Ann Heng$^1$ }

 \address{$^1$ Department of Computer Science and Engineering, The Chinese University of Hong Kong, HK \\
     $^2$Department of Computing, Imeprial College London, UK\\
     $^3$ Imsight Medical Technology Inc, China}
\begin{document}
%
\maketitle
\begin{abstract}
Deep neural networks (DNNs) have achieved great success in a wide variety of medical image analysis tasks. However, these achievements indispensably rely on the accurately-annotated datasets. If with the noisy-labeled images, the training procedure will immediately encounter difficulties, leading to a suboptimal classifier. This problem is even more crucial in the medical field, given that the annotation quality requires great expertise. In this paper, we propose an effective iterative learning framework for noisy-labeled medical image classification, to combat the lacking of high quality annotated medical data. Specifically, an online uncertainty sample mining method is proposed to eliminate the disturbance from noisy-labeled images. Next, we design a sample re-weighting strategy to preserve the usefulness of correctly-labeled hard samples. Our proposed method is validated on skin lesion classification task, and achieved very promising results. 

\end{abstract}
\vspace{-1mm}
\section{Introduction}
\begin{figure}[t!]
  \centering
  \centerline{\includegraphics[width=8.5cm]{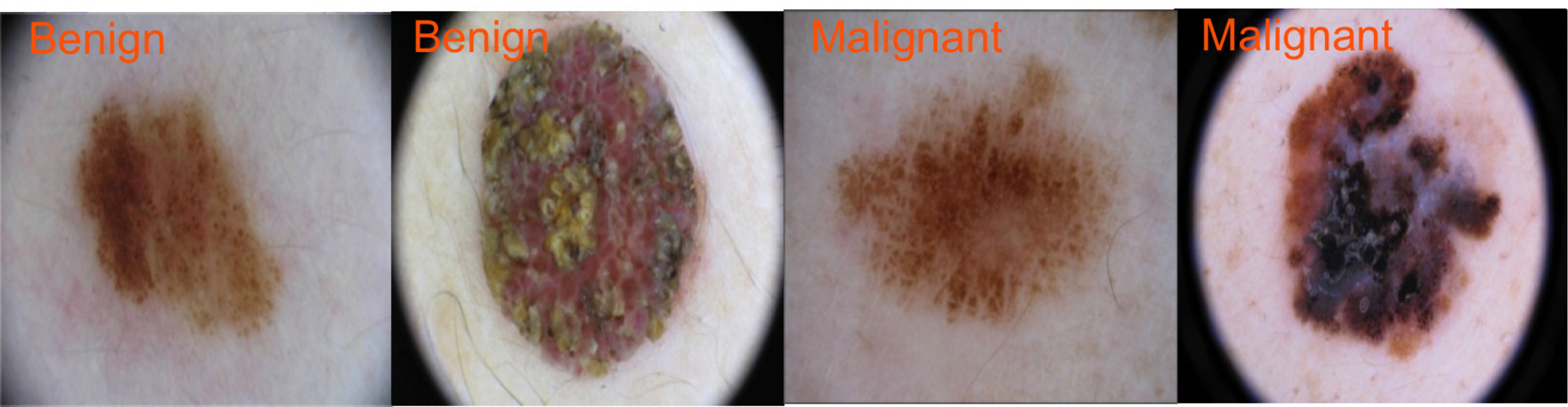}}
\caption{Typical example of melanoma dermoscopy with clinical diagnosis. }
\label{fig:skin}
\vspace{-8mm}
\end{figure}

\vspace{-1mm}
Aiming to improve the performance of Deep Neural Networks (DNNs) on medical image analysis, the community is in the requirement of a huge amount of annotated image data. Meanwhile, the huge capacity of DNNs makes it easily fit noisy labels. Incorrect in training labels can hurt the performance of DNNs on the test dataset \cite{zhang2016understanding}. 
Medical images'annotation quality is prone to experience, which requires years of professional training and domain knowledge. For example, melanoma is the leading death cause of skin cancer, the accuracy of melanoma dermoscopy diagnosis in clinical is 50\% to 82.3\%; the unreliable image label issue can be very severe. With the high demanding of computer-aided diagnosis of melanoma in clinical, it is of significant impact to address the noisy label issue. Despite the label quality problem, DNNs are prone to other training set biases, especially class imbalance and hard samples \cite{Dou2015}. An example of the typical hard samples in melanoma dermoscopy data is shown in Fig.~\ref{fig:skin}, the appearance of benign and malignant cases can be vary similar. These hard samples are normally ambiguous and hence brings about extra challenges for identifying wrong-labelled samples. In this study, we mainly focus on the noisy label issue, as the class imbalance issue can be solved easily during preprocessing or data collection.
\begin{figure*}[ht]
  \centering
  \centerline{\includegraphics[width=17cm]{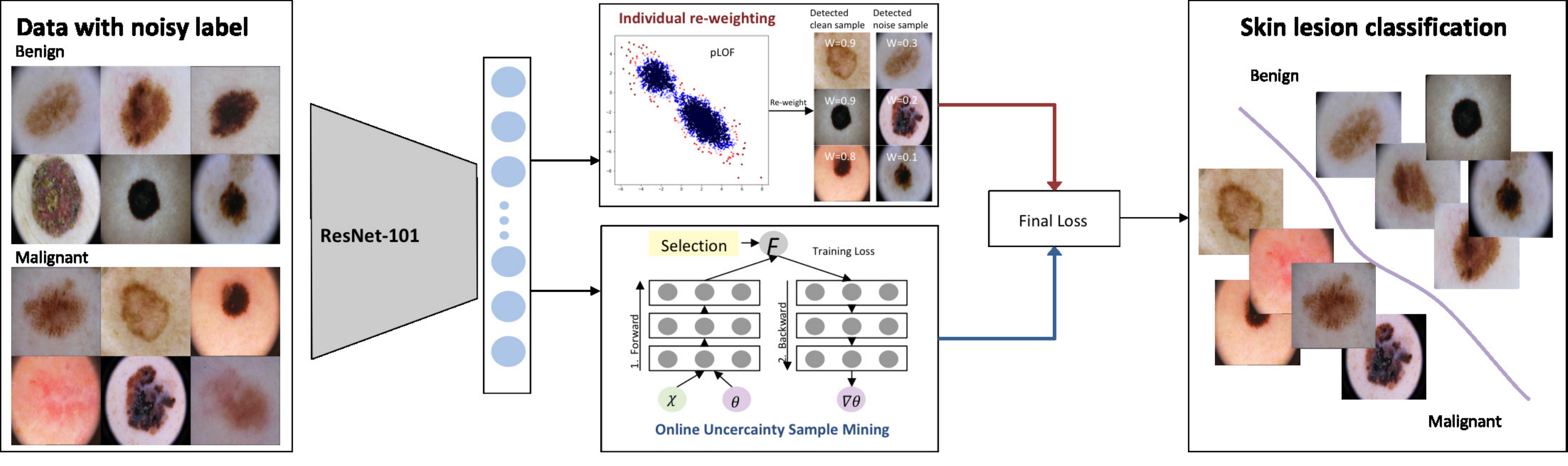}}
\caption{The framework of the proposed learning approach.The network is jointly optimized by two types of losses: re-weighted softmax loss and online uncertainty sample mining loss.}
\label{fig:fram}
\vspace{-4mm}
\end{figure*}

Although some approaches have been considered to address the noisy label issue, it is still an ongoing challenge in deep learning for medical imaging. Aiming to simulate the relationship between noisy label and the latent clean label, Goldberger et al. \cite{goldberger2016training} proposed to add a fully connected layer after softmax, where the updated weight represents the transition matrix between noisy and clean label.  Patrini et al. \cite{patrini2017making} proposed a corrected loss by combining the noise transition matrix with traditional softmax cross entropy loss. These methods are heavily dependent on the accurate assumption of noise distribution, which is usually unknown in real practice. From the assumption that clean data will have a smaller loss than noisy data, Jiang et al. \cite{jiang2017mentornet} proposed Mentornet, which learns small loss samples first. Tanaka et al. \cite{tanaka2018joint} proposed to change the label of training data according to the softmax output during training. Their methods have treated weak agreement sample as noise, but the performance of these methods on medical image are degraded because of the hard samples that are usually presented in the medical image. There are also methods that are supervised by an extra group of clean data, such as a label clean network proposed by Veit et al. \cite{veit2017learning} and an adaptive weight learning method demonstrated by Ren et al. \cite{ren2018learning}. But these methods still need to maintain a set of expert annotated images.

Those methods have demonstrated promising performance in natural images.
Not many studies have addressed the medical image noisy label issue. One pioneer work is \cite{dgani2018training} by Dgani et al., they utilized the method of \cite{goldberger2016training} on mammography classification task and outperforms standard training methods, but it is heavily dependent on the noise label distribution assumption, and the hard samples and minority class in melanoma dataset will obstruct the assumption process.

In this paper, we propose an iterative learning strategy with the aim of detecting noisy label in the training data and enhance the performance of the neural network. Notably, it is a tailor-made strategy for medical images. Specifically, an online uncertainty sample mining strategy is proposed to suppress the noisy samples, and an individual re-weighting module is developed to preserve the hard samples and minority class. Extensive ablation studies demonstrate that the two components both contribute to the performance gain. The main contributions of this paper include: 1) An deep learning model based noisy label training strategy is proposed, which can enhance the model performance when the training data contains noisy labels; 2) A novel noisy label training loss is derived, which considers the hard samples as well as noisy labels.

\vspace{-3mm}
\section{Method}
\vspace{-1mm}
We propose an iterative learning framework that gradually detects noisy samples. As illustrated in Fig.~\ref{fig:fram}, our proposed model consists of three major modules: 1) ResNet-101 network as our backbone model, 2) online sample mining based on uncertainty, 3) sample re-weighting based on deep features.
\vspace{-4mm}
\subsection{Online Uncertainty Sample Mining}
\vspace{-1mm}
The aim of our proposed network is to detect the noisy labeled training data on-the-fly thus reduce the impact of those noisy labels. Considering that during the training process, samples with smaller training losses are more likely to be clean samples, we train the network using those easy samples first. In this regard, an online uncertainty sample mining method (OUSM) is proposed, where only selected samples can contribute to the backpropagation process. During the training process, for each mini-batch, we choose the samples with smaller training loss for backpropagation. In other words, the network is trained by low uncertainty samples as opposed to hard negative mining \cite{dou2017automated}.

More specifically, the online uncertainty sample mining algorithm proceeds as follows. For an input mini-batch with N samples at stochastic gradient descent (SGD) iteration t, we first compute the loss in the forward pass, the loss represents the uncertainty of the current network on the sample. Then, the high uncertainty samples of the batch are selected by sorting the input images according to training loss and taking the top K examples out of the N samples. The loss of the selected K samples are set to be 0, and hence no gradient update for these samples. The whole training process is more effective as only part of samples are selected for updating the model, the backward pass is not as expensive as before.
\vspace{-4mm}
\subsection{Individual Re-weighting Loss}
\vspace{-1mm}
For medical images, the imbalance distribution and existence of hard examples are very common. As we select the low uncertainty samples to do backpropagation during training, it is likely that the minority class and hard samples are ignored. Therefore, we propose a re-weighting loss to preserve the influence of the minority class and the hard samples. For each sample, we assign a weight (between 0 and 1) individually based on how likely one sample being noisy, the detected noisy samples will have smaller weights, and on the contrary, the detected clean samples will have larger weights. The individual weight is calculated based on the pre-softmax layer feature using a probabilistic Local Outlier Factor algorithm (pLOF) cumulatively, the score is scaled to a probabilistic value in [0,1] similar to \cite{Kriegel}. Local Outlier Factor (LOF) is an unsupervised algorithm to detect outlier on high dimensional data, the intuition is the density around an outlier object is significantly different from the density around its neighbors. The pLOF score can be directly interpreted as the probability of a sample being an outlier, in our setting, it is the probability of one being a noisy sample. We used 1-pLOF to weight each sample. 
The initial weight of each sample is assigned as 1, and the training samples are reweighted every five epochs. 

Overall, the network is jointly optimized by the following loss: 
\begin{equation} \label{eq1}
\mathcal{L}=\alpha\times\mathcal{L}_{OUSM}+\beta\times\lambda\times\mathcal{L}_{CE}+\eta\Vert W \Vert^{2}
\end{equation}
where  $\mathcal{L}_{OUSM}$ is the online uncertainty sample mining loss, where only N-K samples contribute to the backpropagation process, and the reweighting factor $\lambda=1-pLOF$, $\mathcal{L}_{CE}$ is the cross entropy loss.  $\eta$ denotes the weight decay term and $W$ denotes the parameters of the whole network. $\alpha$ and $\beta$ are the trade-off parameters between these terms. In this study, we set both $\alpha$ and $\beta$ equal to 0.5.
\section{Experiments and Results }
\vspace{-1mm}
\subsection{Data and Preprocessing}
We justify the performance of our proposed framework on skin lesion analysis challenge ISIC 2017 Challenge Part 3: Skin Lesion Classification. The dataset consists of 2,000 dermoscopy images for training, and we downloaded additional 1,582 images from the ISIC archive\cite{codella2018skin} to enhance the training data. In total, we have 3,582 training images, which consists of 2,733 benign samples and 894 malignant samples. We use the original 150 validation datasets and 600 test datasets as our validation data and test data. All the images are resized to $224 \times 224$. 
We pre-processed the data by subtracting the mean value of Imagenet before inputting into the network. To simulate the clinical wrong diagnosis,  a fine-tuned ResNet-101 classifier was trained on random selected 1,000 images and tested on the other 2,582 images, and the 2,582 images was sorted according to the test loss. The top $\gamma$ $(\gamma \in{0.05, 0.1 ,0.2, 0.4})$ percent high loss samples in each class (2,733 benign and 894 malignant) were selected symmetrically as noisy label. The noisy label is defined as $y^{\prime}_{i}=y_{i} $ with the probability of $ 1-\gamma$, and $y^{\prime}_{i}=y_{k},y_{k}\neq y_{i} $ with the probability of $\gamma$, where $y^{\prime}_{i}$ is the corrupted noisy label,$y_{i}$ is the clean label. Our noisy label learning network was implemented with Keras with Tensorflow backend. 
\vspace{-3mm}
\subsection{ Memorization of Neural Networks to Noisy Data}
\begin{figure}[t]
\begin{minipage}[b]{.48\linewidth}
  \centering
  \centerline{\includegraphics[width=4.0cm]{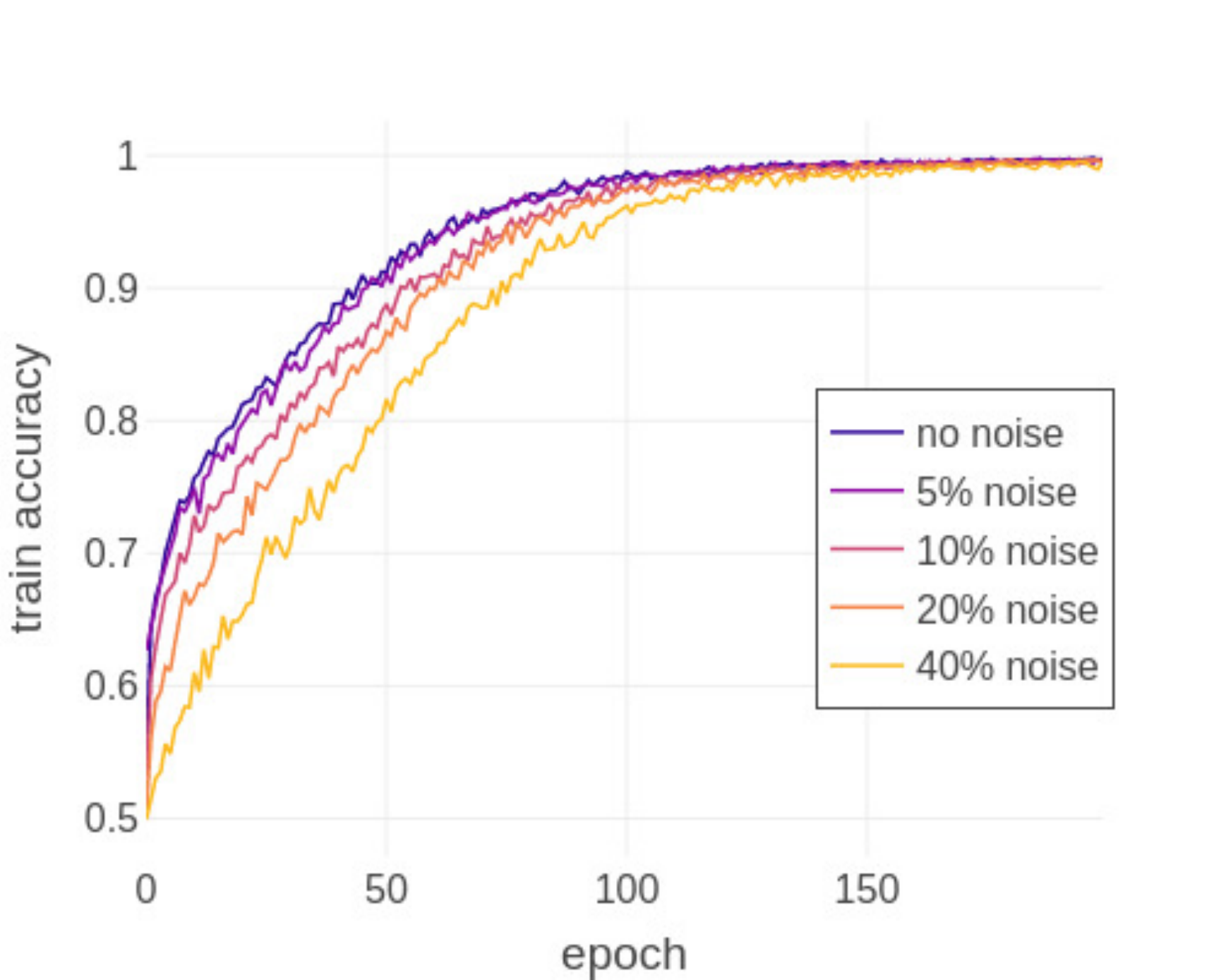}}
\end{minipage}
\hfill
\begin{minipage}[t]{0.48\linewidth}
  \centering
  \centerline{\includegraphics[width=4.0cm]{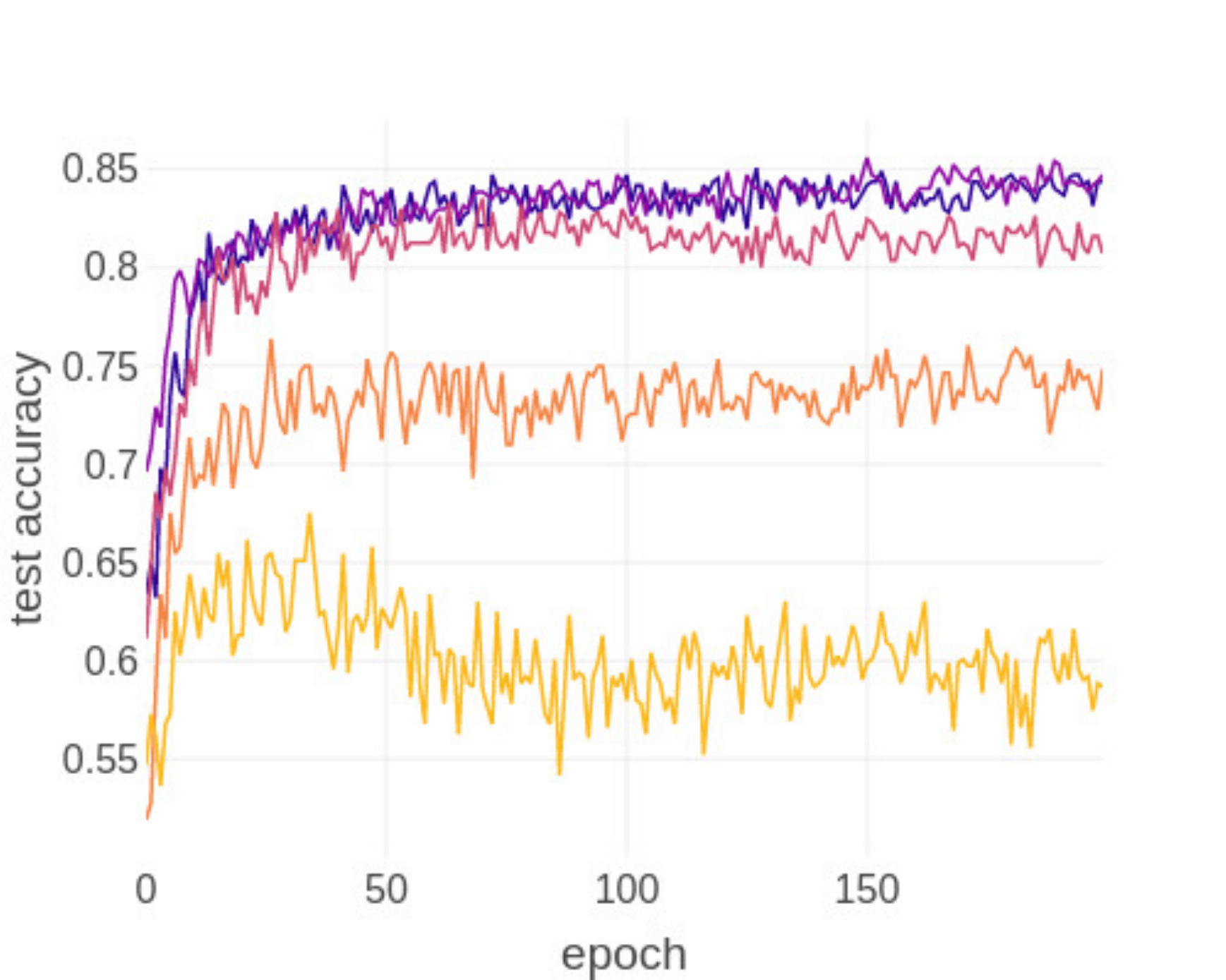}}
  \vspace{-1.5cm}
\end{minipage}
\caption{Skin lesion classification performance with corrupted noisy label. Left: Training curve of the neural network with different noisy corruption percentage (5\%, 10\%, 20\%, 40\%); Right: The corresponding test curve of the neural network on the official test data with different noisy corruption percentage (5\%, 10\%, 20\%, 40\%).}
\vspace{-0.5cm}
\label{fig:capacity}
\end{figure}

\vspace{-1mm}
To evaluate the memorization of deep learning to noisy labeled training dataset. We firstly trained ResNet-101 network initialized with Imagenet weight on the aforementioned corrupted noisy datasets with different noisy ratio $\gamma$, where $\gamma \in (0.05, 0.1, 0.2, 0.4)$. The network was trained with SGD, and  a learning rate of 1e-4, weight decay 1e-4, the batch-size is 32. Fig.~\ref{fig:capacity}. shows the learning curves of the network on the selected noisy labeled datasets under various noisy components  $\gamma$. From the figure, we can observe the neural network can easily fit the noisy label in the training dataset. For example, the test error dramatically dropped with increased noisy components. However, the training error always converged to 0. The phenomenon validates the hypothesis that the DNNs can overfit to the noisy annotated medical images, it can use brute-force memorization to fit the noise in medical image label.
From Fig.~\ref{fig:capacity}. we also obtain that clean data are easier to fit than noisy data, with the increasing of the noise percentage, DNNs need more epochs to fit all the training data. We also obtain that, when the training data contains noisy labels, the test accuracy of validation data will decrease through the training as DNNs learn simple patterns first before memorizing the noise \cite{Arpit2017}. Therefore, we propose to first train the network without individual re-weighting, where the individual weight of all the samples is set to be 1 in the first five epochs.
\vspace{-3mm}
\subsection{Quantitative Evaluation and Comparison}
\vspace{-1mm}
We extensively validated our method on the symmetric noisy samples. For all the experiments, we employed the ResNet-101 pretrained on Imagenet as our backbone model with batch-size 32. The learning rate is 1e-3 and weight decay is 1e-4. The model was trained with SGD. In this paper, we evaluate different methods using 0.5 threshold Accuracy.
 
Firstly, Our upper bound is acquired by training a ResNet-101 with weight initialized by Imagenet using cross-entropy loss. The accuracy achieved 86.3\%, which is on the third place on the challenge leaderboard. The results indicate the reliability of using ResNet-101 as our backbone. Our lower bound is acquired using cross-entropy loss to train the model on each corrupted datasets, as demonstrated in Table \ref{tab:compare}. To investigate the contribution of each strategy adopted in our method, we perform ablation studies by removing the individual re-weighting module. The model with different loss formulations and ensemble strategies are reported in Table \ref{tab:compare}. By utilizing the online uncertainty sample mining, the accuracy for all the datasets has been improved. By adding the individual re-weighting module, the model gains further performance improvements and achieves the best Accuracy score. The ablation study indicates that only select low uncertainty small loss samples cannot solve the noisy label issue for medical image analysis. The main reasons could be the existence of the hard samples and imbalanced data distribution between the malignant class and the benign class. As shown in Table \ref{tab:compare}. The ensemble of the individual re-weighting module can boosts the performance, as the influence of hard samples and minority class in the dataset is preserved. 

We compared the performance of our method with the state-of-the-art method \cite{patrini2017making}. We utilized the forward loss proposed in \cite{patrini2017making} and estimated the noise transition matrix follow the description of the paper. The evaluation results in Table \ref{tab:compare} highlight that our proposed method outperforms the state-of-the-art method by a significant margin.

\newcolumntype{C}{>{\centering\arraybackslash}X}
\setlength{\extrarowheight}{1pt}
\begin{table}
	\vspace{-2mm}
	\caption{Comparison of noisy labeled skin image classification task results applying different methods. ResNet represents training on selected noisy data using ResNet-101 pretrained on Imagenet; ResNet+OUSM represents using online uncertainty sample mining (OUSM).}
	\centering
	\begin{tabularx}{0.48\textwidth}{@{}l*{5}{C}}
		\toprule
		Noise Ratio          & 0\%  & 5\%   & 10\%   & 20\%   & 40\%    \\ \midrule
		ResNet & 86.3    &82.6     &79.1    & 75.2    & 65.3    \\  \hline
		Panini\cite{patrini2017making}  &--    &83.2    &81.3     &79.7     &68.4    \\ \hline
		ResNet+OUSM    &--    &82.5    &81.2        &77.6    &73.6    \\ 
		Our method  & -- &\textbf{84.5}     &\textbf{83.6}    & \textbf{80.7} & \textbf{75.7}     \\ 
		\bottomrule
	\end{tabularx}
	\label{tab:compare}
	\vspace{-5mm}
\end{table}
\vspace{-2mm}
\section{Conclusion}
\vspace{-2mm}
This paper tackles the label noise caused by wrong diagnosis training data, or the dismission from the radiologists. We present an iterative learning framework for learning on noisy labeled medical images by online uncertainty sample selection and individual sample reweighting. The application to the noisy skin lesion data shows that the accuracy can be improved by a large margin when the training data contains noisy labels. In addition, the advance of our method is that our method can be trained in an end-to-end manner, no pre-estimation of noise distribution or extra clean data is needed. The proposed strategy can be applied in any other classification context.\\
\vspace{-7mm}
\section{Acknowledgement}
\vspace{-2mm}
This project is funded by Hong Kong Innovation and Technology Commission, under ITSP Tier 2 Scheme (Project No. ITS/426/17FP), and ITSP Tier 3 Scheme ( Project No. ITS/041/16)\\

\bibliographystyle{IEEEbib}
\vspace{-8mm}
\bibliography{ref}
\end{document}